\def\cvprPaperID{12215}
\def\confName{ICCV}
\def\confYear{2025}

\def\authorBlock{
    Siyuan Yao$^{1,2}$, Rui Zhu$^{1}$, Ziqi Wang$^{1}$, Wenqi Ren$^{2,4,5}$, Yanyang Yan$^{3}$, Xiaochun Cao$^{2}$\thanks{\rm Corresponding Author.} \\
    $^{1}$ Beijing University of Posts and Telecommunications 
    $^{2}$ Sun Yat-sen University \\
    $^{3}$ University of Chinese Academy of Sciences 
    $^{4}$ MoE Key Laboratory of Information Technology\\
    $^{5}$ Guangdong Key Laboratory of Information Security Technology\\
    \tt\small yaosiyuan04@gmail.com,
    \tt\small ruizhu@bupt.edu.cn,
    \tt\small zq\_wang@bupt.edu.cn, \\
    \tt\small yanyanyang@ict.ac.cn,
    \tt\small rwq.renwenqi@gmail.com, \tt\small caoxiaochun@mail.sysu.edu.cn
}

\newif\ifreview 
\newif\ifarxiv \newcommand{\arxiv}{\arxivtrue}
\newif\ifcamera 
\newif\ifrebuttal 

\arxiv

\pdfoutput=1

\documentclass[10pt,twocolumn,letterpaper]{article}
\usepackage{bm} 
\usepackage{ulem}

\ifreview \usepackage[review]{cvpr} \fi
\ifarxiv \usepackage[pagenumbers]{cvpr} \fi
\ifrebuttal \usepackage[rebuttal]{cvpr} \fi
\ifcamera \usepackage{cvpr} \fi

\usepackage{graphicx}
\usepackage{amsmath}
\usepackage{amssymb}
\usepackage{booktabs}

\usepackage{times}
\usepackage{microtype}
\usepackage{epsfig}
\usepackage[table,xcdraw]{xcolor}
\usepackage{caption}
\usepackage{float}
\usepackage{placeins}
\usepackage{color, colortbl}
\usepackage{stfloats}
\usepackage{enumitem}
\usepackage{tabularx}
\usepackage{xstring}
\usepackage{multirow}
\usepackage{xspace}
\usepackage{url}
\usepackage{subcaption}
\usepackage{xcolor}
\usepackage[hang,flushmargin]{footmisc}
\usepackage{booktabs}

\ifcamera \usepackage[accsupp]{axessibility} \fi

% ...

\ifarxiv  \fi

\newcommand{\R}[1]{{%
    \textbf{%
        \ifstrequal{#1}{1}{\textcolor{red}{R#1}}{%
        \ifstrequal{#1}{2}{\textcolor{blue}{R#1}}{%
        \ifstrequal{#1}{3}{\textcolor{magenta}{R#1}}{%
        \ifstrequal{#1}{4}{\textcolor{teal}{R#1}}{%
                           \textcolor{cyan}{R#1}%
        }}}}%
    }%
}}

\usepackage{xr-hyper}

\makeatletter
\newcommand*{\addFileDependency}[1]{
  \typeout{(#1)}
  \@addtofilelist{#1}
  \IfFileExists{#1}{}{\typeout{No file #1.}}
}

\makeatother

\usepackage[pagebackref,breaklinks,colorlinks]{hyperref}
\usepackage[capitalize]{cleveref}
\crefname{section}{Sec.}{Secs.}
\crefname{table}{Table}{Tables}
\crefname{figure}{Fig.}{Figs.}

\usepackage{cancel}
\begin{document}
\title{UMDATrack: Unified Multi-Domain Adaptive Tracking \\
Under Adverse Weather Conditions}
\author{\authorBlock}
\maketitle

\begin{abstract}
Visual object tracking has gained promising progress in past decades. Most of the existing approaches focus on learning target representation in well-conditioned daytime data, while for the unconstrained real-world scenarios with adverse weather conditions, e.g. nighttime or foggy environment, the tremendous domain shift leads to significant performance degradation. In this paper, we propose UMDATrack, which is capable of maintaining high-quality target state prediction under various adverse weather conditions within a unified domain adaptation framework. Specifically, we first use a controllable scenario generator to synthesize a small amount of unlabeled videos (less than $2\%$ frames in source daytime datasets) in multiple weather conditions under the guidance of different text prompts. Afterwards, we design a simple yet effective domain-customized adapter (DCA), allowing the target objects' representation to rapidly adapt to various weather conditions without redundant model updating. Furthermore, to enhance the localization consistency between source and target domains, we propose a target-aware confidence alignment module (TCA) following optimal transport theorem. Extensive experiments demonstrate that UMDATrack can surpass existing advanced visual trackers and lead new state-of-the-art performance by a significant margin. Our code is available at \href{https://github.com/Z-Z188/UMDATrack}{https://github.com/Z-Z188/UMDATrack.}
\end{abstract}
\section{Introduction}
\label{sec:intro}

Visual object tracking (VOT) is a fundamental visual task of computer vision over the past decades, aiming to estimate the state of arbitrary target objects in video sequences given the initial annotation. Existing mainstream methods formulate object tracking as a target matching problem, which constructs template-search pairs to learn a position-sensitive matching network for target localization. Owing to the promising advances of recent deep learning architectures, VOT has achieved remarkable success in terms of accuracy and efficiency.

\begin{figure}[!t]
\centering

\includegraphics[ width=3.4in]{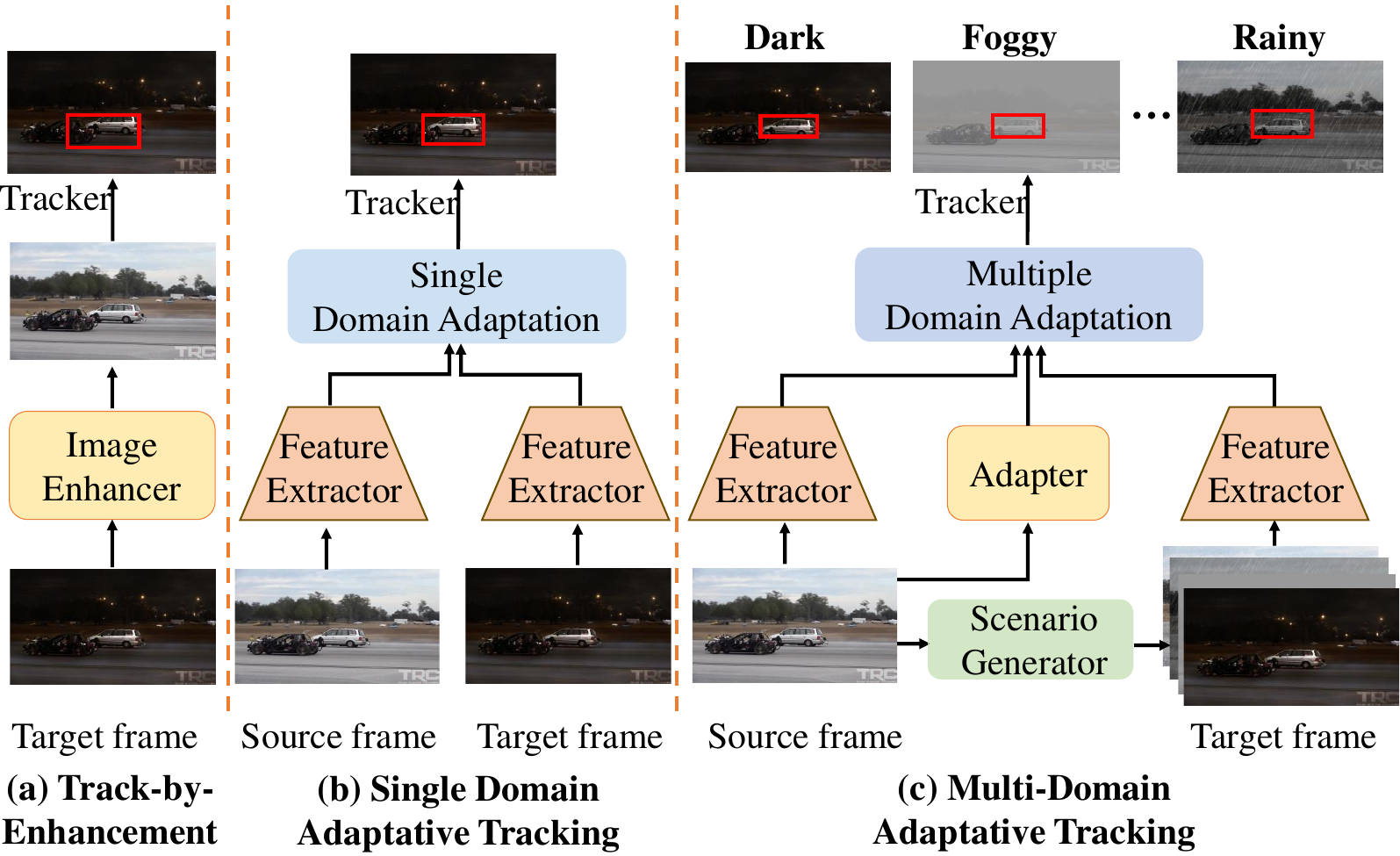}
\caption{Three representative tracking pipelines under adverse weather conditions. (a) "Track-by-Enhancement" pipeline \cite{DBLP:conf/iros/Ye0ZCL21}. (b) Single domain adaptation pipeline \cite{DBLP:conf/cvpr/Ye0ZPC22}. (c) The proposed unified multi-domain adaptive tracking (UMDATrack) pipeline. UMDATrack utilizes controllable scenarios generator to synthesize unlabeled video frames and employ a flexible domain-customized adapter to transfer the knowledge to multi-domain.}
\label{fig:motivation}
\vspace{-10pt}
\end{figure}

Recent advanced object trackers typically utilize well-conditioned daytime datasets, e.g. LaSOT \cite{DBLP:conf/cvpr/FanLYCDYBXLL19} or TrackingNet \cite{DBLP:conf/eccv/MullerBGAG18} as supervision for model training, however, the performance of these SOTA trackers is unsatisfactory in real-world scenarios with adverse weather conditions (e.g. nighttime or foggy environment) due to the tremendous domain gap. To address this issue, some efforts have explored to introduce synthesized datasets \cite{zhang2021domain,DBLP:conf/aaai/WuYHWZZR25} or domain adaptive discriminator \cite{DBLP:conf/cvpr/Ye0ZPC22,DBLP:conf/icra/ZhuTCH0QLL24} to enhance the cross-domain transferability. Despite the significant advances, they potentially suffer from two drawbacks. First, most of the existing approaches are designed for single weather condition, while the generalization abilities are greatly limited in various scenarios where multiple target domains are available. For example, as shown in Fig. \ref{fig:motivation}, the nighttime tracker UDAT \cite{DBLP:conf/cvpr/Ye0ZPC22} is capable of predicting the target state in nighttime data, but its performance drops significantly when the environment changed to another foggy weather condition. Besides, recent domain adaptive trackers generate large amounts of target domain samples for model knowledge transfer, the sample generation process is time-consuming and the intrinsic relationship of the target objects in multiple domains has been overlooked. For different weather conditions in multiple target domains, existing approaches require to introduce redundant parameters to conduct feature alignment separately, which fails to perform cross-domain interaction in an efficient manner.

In this paper, we propose a unified multi-domain adaptive tracker termed UMDATrack, which is capable of maintaining high-quality target state prediction under various adverse weather conditions.  Inspired by the great success of the controllable text-to-image generation technique, we first utilize a text-conditioned diffusion model to synthesize unlabeled videos in multiple weather conditions under the guidance of different text prompts. Afterwards, to flexibly transfer the target objects' representation from source domain to multiple target domains, we froze the backbone feature extractor and design a simple yet effective domain-customized adapter (DCA) to remedy the tracking model, allowing it to be rapidly adapted to various weather conditions without redundant model updating. Furthermore, we propose an target-aware confidence alignment module (TCA) with optimal transport theorem, which enhances the localization consistency between source and target domains by measuring the discrepancies of the localization confidence at the candidate positions. Experiments show that by only synthesizing a small partition of videos (less than $2\%$ frames in source domain) at arbitrary weather conditions, UMDATrack can surpass existing advanced visual trackers and lead new state-of-the-art performance on either real-world or synthesized datasets by a significant margin. \emph{To the best of our knowledge, this is the first unfiied multi-domain adaptation tracker in VOT community}. 

In summary, the main contributions of this work can be concluded in three aspects:

$\bullet$  We propose a unified multi-domain adaptive tracking framework termed UMDATrack, which conducts multi-domain transfer using text-conditioned diffusion model and maintains high-quality target state prediction under various adverse weather conditions.

$\bullet$ We design a simple yet effective domain-specific adapter (DCA) to remedy the tracking model, which can flexibly transfer the target objects' representation from original daytime scenario to various weather conditions without redundant model updating.

$\bullet$ We propose a target-aware confidence alignment module (TCA) with optimal transport theorem to enhance the localization consistency in source and target domains. Extensive experiments demonstrate that UMDATrack achieves superior performance to existing state-of-the-art methods.

\section{Related Work}
\label{sec:related}

\subsection{Tracking in Adverse Weather Conditions}
Recently, object tracking in adverse weather conditions has attracted increasing interest due to a variety of practical applications. The classical methods employ multi-modal sensors, e.g. Visible+Depth (RGB-D) \cite{DBLP:conf/iccv/YanYKZLK21}
Visible+Thermal (RGB-T) \cite{DBLP:conf/cvpr/WangXCZZZY20} for target appearance modeling in complex scenarios. However, these methods require to collect large amount of labelled examples to learn the cross-modal target representation. To address this issue, some works explore to use the RGB images only to transfer the knowledge to unlabelled target domains. Existing methods generally \cite{zhang2021domain,DBLP:conf/iros/Ye0ZCL21} perform image enhancement to unify target object's representation. For example, Zhang \emph{et~al.} \cite{zhang2021domain} combine RGB images and the corresponding depth maps to synthesize the foggy images. The feature alignment is conducted on Siamese trackers \cite{DBLP:conf/cvpr/ChenZLZJ20,yao2021robust,yao2021learning} using the synthesized foggy datasets to eliminate the semantic-level domain shift. HighlightNet \cite{DBLP:conf/iros/0001DY0LZ22} adapts to illumination variation and excavates the potential object for low-light UAV tracking. UDAT \cite{DBLP:conf/cvpr/Ye0ZPC22} proposes a transformer-based bridging layer to transfer the semantic knowledge from daytime domain to the nighttime domain. Though effective, the aforementioned trackers are designed for single weather condition, while the generalization abilities are greatly limited in various weather conditions where multiple target domains are available.

\subsection{Controllable Text-to-Image Generation}

To transfer the knowledge in various weather conditions, the scene translation technique has been introduced to synthesize high-quality images. The early efforts use Generative Adversarial Networks (GANs) \cite{DBLP:conf/cvpr/IsolaZZE17} to transform images from source domain to target domain by modifying image style. However, these GAN-based methods typically require training from scratch on the specific domains. Recently, the advanced text-to-image (T2I) diffusion models \cite{DBLP:conf/iccv/GePZ023,DBLP:conf/iccv/ZhangRA23} have shown impressive controllable flexibilities using text descriptions. GLIDE \cite{DBLP:conf/icml/NicholDRSMMSC22}
trains a CLIP model in noisy image space to provide CLIP guidance for image generation and editing. DALL-E \cite{DBLP:conf/icml/RameshPGGVRCS21} employs an autoregressive transformer to combine both text and image tokens, which demonstrates remarkable zero-shot translation capabilities without using large-scale training samples. ControlNet \cite{DBLP:conf/iccv/ZhangRA23} treats the pretrained model as a strong backbone and finetune the trainable copy connected with zero convolution layers, allowing users to add various spatial conditions to control the image generation. Inspired by the success of these text-to-image (T2I) generation models, in this work, we utilize text-conditioned diffusion model to synthesize unlabeled videos in multiple weather conditions for target feature translation.
\begin{figure*}
  \centering
  \includegraphics[width=7in]{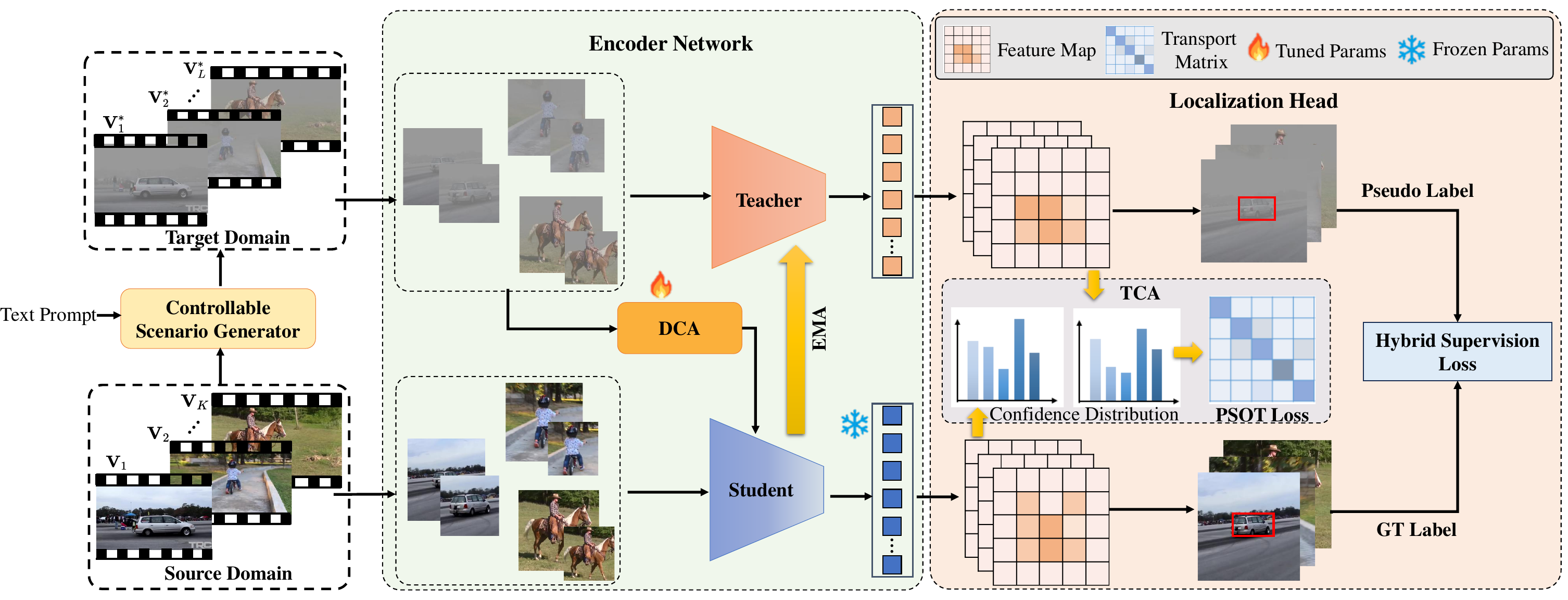}
  \caption{ Overview of the proposed UMDATrack. It first utilizes a controllable scenarios generator (CSG) to synthesize the video frames in arbitrary adverse weather conditions. The cropped template-candidate pairs are sent into a student-teacher network, which transfers the target objects’ representation to multiple weather conditions using an encoder network with domain customized adapter (DCA) and a localization head with target-aware confidence alignment module (TCA). Here we only demonstrate the \textit{daytime} $\to$ \textit{foggy} environment translation for simplicity.
  }
  \label{fig:framework}
\end{figure*}

\subsection{Multi-Target Domain Adaptation}
Recently, various techniques have been employed for Multi-Target Domain Adaptation (MTDA) to enhance cross-domain robustness and generalization. For example, curriculum learning and feature aggregation have been combined to align similar features and adapt models gradually to domain complexities \cite{DBLP:conf/cvpr/RoyKZS021}. Other approaches\cite{li2024training} have explored merging independently adapted models from distinct domains by combining model parameters and buffer merging. Additionally, graph matching techniques \cite{sigma} have been applied to improve generalization in cross-domain object detection, with self-training methods also showing promising potential. Optimal transport theory has been widely studied and applied across various domains. A regularized unsupervised optimal transport model\cite{7586038} has been proposed to align source and target domain representations, using a transport plan that enhances cross-domain robustness. In particular, SOOD \cite{DBLP:conf/cvpr/HuaLLLZYB23} uses optimal transport to ensure global layout consistency between pseudo-labels and predictions. Despite the aforementioned efforts, it is still challenging to design a unified tracker to conduct MTDA in adverse weather conditions like fog, nighttime, and rain. Our research effectively fills this gap by leveraging optimal transport theory to improve tracking robustness in these challenging scenarios.

\section{Method}
\label{sec:method}

In this section, we describe the overall architecture of the proposed UMDATrack, which consists of three main components: a controllable scenarios generator (CSG), an encoder network with domain customized adapter (DCA) and a localization head with target-aware confidence alignment module (TCA).

\subsection{Controllable Scenario Generator}
\begin{figure}[!t]
  \centering
  \includegraphics[width=3.4in]{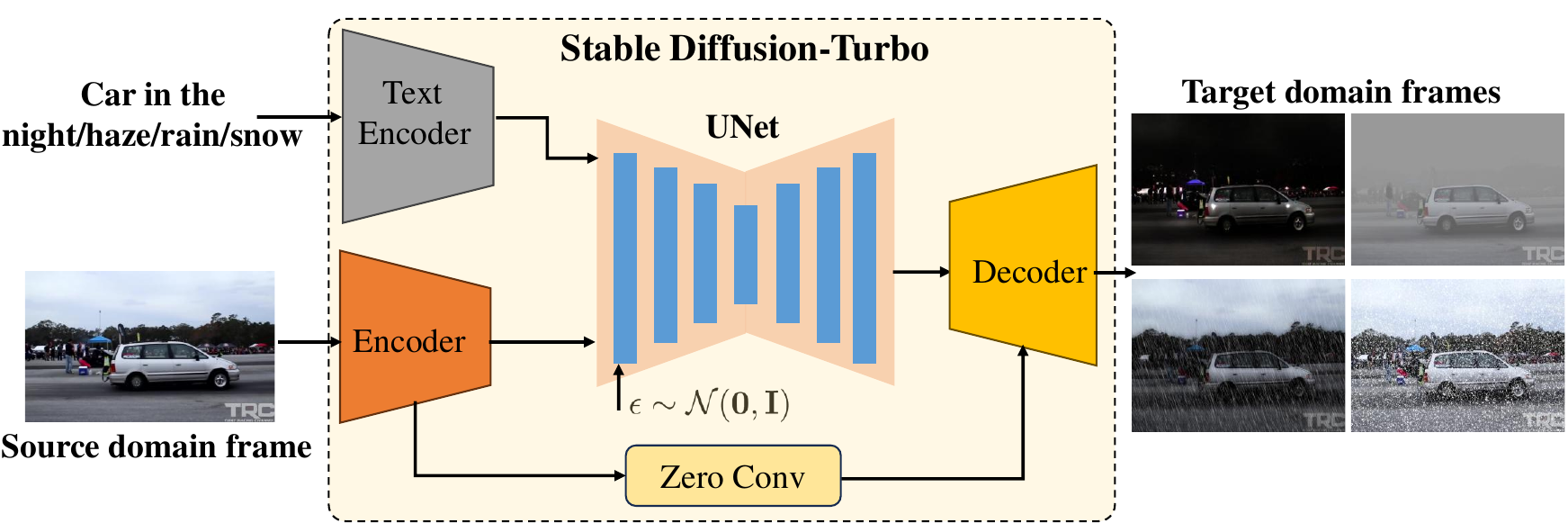}
  \caption{Details of the Controllable Scenario Generation (CSG) module.}
  \label{fig:CSG}
\end{figure}

As it is not trivial to collect large number of video sequences in adverse weather conditions, we first synthesize a small amount of training data to conduct domain knowledge transfer. Inspired by recent advances of text-to-image (T2I) techniques, we utilize a controllable scenario generator (CSG) for data synthesis. Let $\mathbb{V}=\left \{\mathbf{V}_{1}, \mathbf{V}_{2},\cdots,\mathbf{V}_{K}  \right \}$ denotes the videos in source domain and $\mathbb{V}^{\ast }=\left \{\mathbf{V}^{\ast }_{1}, \mathbf{V}^{\ast }_{2},\cdots,\mathbf{V}^{\ast }_{L}  \right \}$ denotes the videos in target domain, here $L\ll K$ indicates the size of $\mathbb{V}^{\ast }$ is significantly smaller compared to $\mathbb{V}$. Our goal is to randomly select the videos in $\mathbb{V}$ and translate them to arbitrary weather conditions, e.g. hazy, dark and rainy, etc. To achieve this, we use the T2I model, \emph{i.e. Stable Diffusion-Turbo} \cite{DBLP:conf/eccv/SauerLBR24} to translate the scenarios using different text prompts. As shown in Fig. \ref{fig:CSG}, the text prompt $c_X$, \emph{e.g.} \textit{"Car in the night/haze/rain/snow"} and the video images $x \in \mathbb{V}$ in source domain are fed into the text encoder and image encoder respectively. We generate the output video frames $y\in \mathbb{V}^{\ast }$ in target domain by integrating video frame $x$ with conditional controls $c_{X}$ and the noise $\epsilon$ as:
\begin{equation}
	\label{generator}
     y =G_{\mathrm{SDT}}(x,c_{X},\epsilon ), 
     \epsilon \sim \mathcal{N}(\mathbf{0} ,\mathbf{I}),
\end{equation}
where $G_{\mathrm{SDT}}(x, c_{X}, \epsilon)$ denotes the Stable Diffusion-Turbo generator, $\epsilon$ is the noise map. The skip connections and Zero-Convs are used to preserve the essential structural details of the images. Benefited from the powerful transferability of T2I model, the video frames in target domains can be rapidly generated within only 1-4 iteration steps by simply changing the text prompts.


\subsection{Tracking in Multiple Weather Conditions}
Though CSG can generate continuous video frames in multiple weather conditions, the appearance discrepancies of target objects between the source daytime videos and the synthesized videos still limit the tracker's generalization ability. To address this issue, we design a unified domain adaptation framework following the teacher-student pipeline, which can be flexibly deployed to various domain-customized scenarios. Specifically, given $N_{\mathcal{S}}$ video frames $\mathcal{D}_{\mathcal{S}} = \{(\mathcal{I}_i^{\mathcal{S}}, \mathbf{b}_i^{\mathcal{S}})\}_{i=1}^{N_{\mathcal{S}}}$ in source domain and $N_{\mathcal{T}}$ unlabeled frames $\mathcal{D}_{\mathcal{T}} = \{\mathcal{I}_{i}^{\mathcal{T}} \}_{i=1}^{N_{\mathcal{T}}}$, where $\mathcal{I}_i^{\mathcal{S}}$ and $\mathbf{b}_{i}^S$ denotes the images and annotated bounding boxes in the source domain, $\mathcal{I}_{i}^{\mathcal{T}}$ denotes the images in multiple target domains. We crop the paired template-search images of $\mathcal{D}_{\mathcal{S}}$  and $\mathcal{D}_{\mathcal{T}}$ and then send them into the student and teacher network, respectively. 
The \textbf{student} $\to$ \textbf{teacher} knowledge transfer is conducted by updating the weights of the teacher model using the EMA (Exponential Moving Average) as:

\begin{equation}
	\label{attn2}
    \theta^{\mathcal{T}} \gets \alpha \theta^{\mathcal{T}}+(1-\alpha) \theta^{\mathcal{S}},
\end{equation}
where $\theta^{\mathcal{T}}$ and $\theta^{\mathcal{S}}$ denote the learnable parameters of the teacher and student networks. $\alpha$ is the momentum coefficient controlling the updating rate of the teacher.


\noindent\textbf{Domain-Customized Adapter} The student-teacher training paradigm allows the tracker to gradually propagate source domain information to target domain. However, as the data distributions in different weather conditions vary greatly, it's time-consuming to generate large amounts of multi-domain samples and would inevitably introduce redundant parameters if we conduct domain knowledge transfer separately. Considering this, we propose a Domain Customized Adapter (DCA) to transfer the target object's representation to arbitrary weather conditions in an efficient fashion.

\begin{figure}[!t]
  \centering
  \includegraphics[width=3.3in]{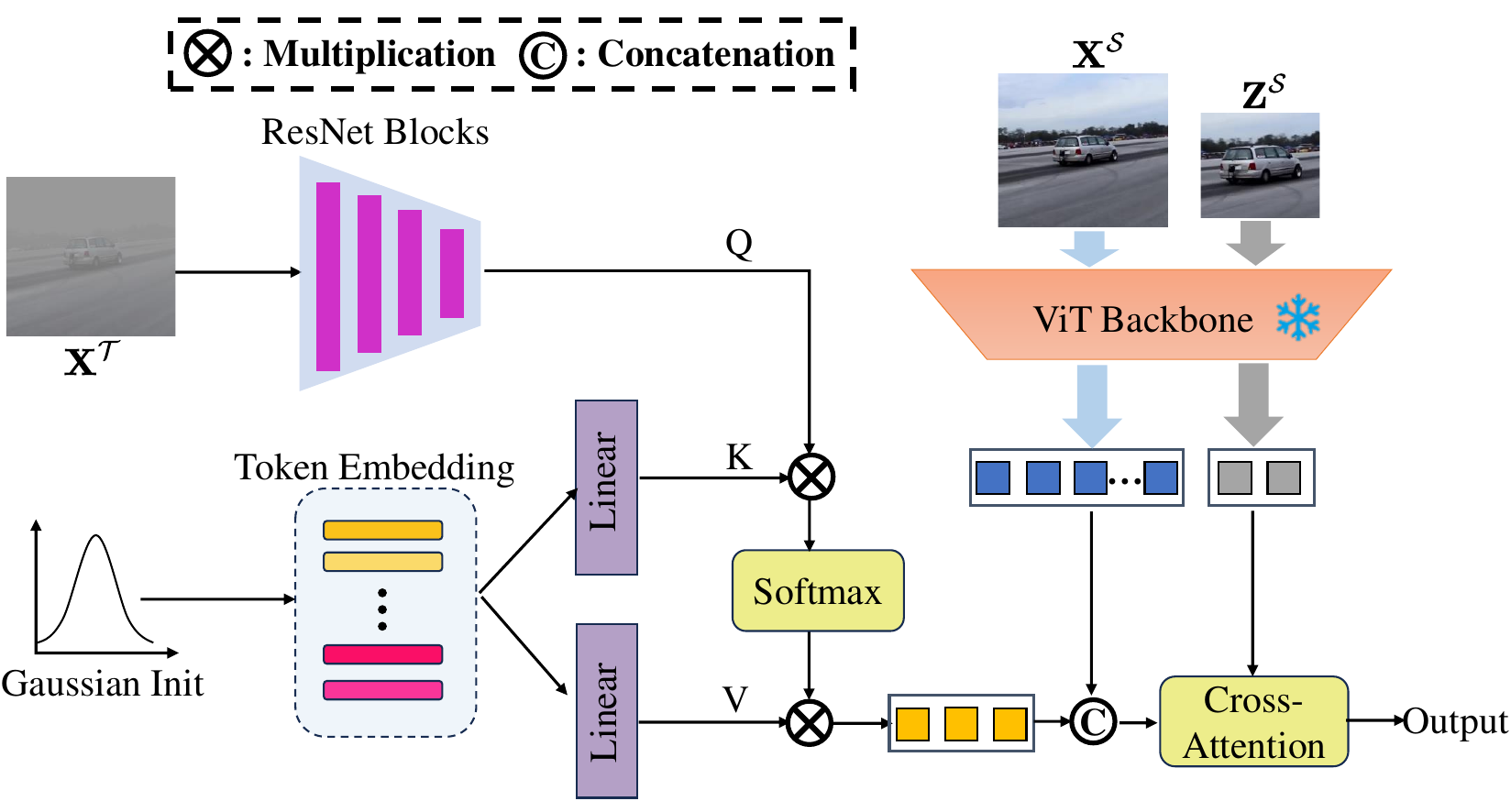}
  \caption{Details of the Domain-Customized Adapter (DCA) module.}
  \label{fig:adapter}
\end{figure}

We present the detailed structure of DCA in Fig. \ref{fig:adapter}. Formally, suppose the cropped template-search images in source domain are $\mathbf{Z}^{\mathcal{S}}$ and $\mathbf{X}^{\mathcal{S}}$, respectively. While the  image pairs in target domain are $\mathbf{Z}^{\mathcal{T}}$ and $\mathbf{X}^{\mathcal{T}}$. We first use a lightweight ResNet block to transform and reshape $\mathbf{X}^{\mathcal{T}}$ as query $\mathbf{Q} \in \mathbb{R}^{K \times C}$. Then we initialize a Gaussian random variable and embed it to be learnable token bank $\mathbf{B} \in \mathbb{R}^{L' \times C}$ that consists of $ L'$ learnable feature vectors with channel dimension \( C \). The token bank $B$ is further projected as key-value tokens $K$ and $V$ with the size of $L' \times C$ by two FC layers, respectively. We compute an structural token $\mathbf{S}$ between the query and embedded key-value tokens as follows:

\begin{equation}
	\label{cross-attention}
		 \mathbf{S}= {\mathrm{Softmax}}(\frac{\mathbf{Q}\mathbf{K}^\top}{\sqrt{d_k}})\mathbf{V},
\end{equation}
the structural token $\mathbf{S} \in \mathbb{R}^{ K \times C }$ encodes the latent image content representation, which shares similar contextual structure to $\mathbf{X}^{\mathcal{S}}$ in the embedding space. The structural token $\mathbf{S}$ are subsequently fed into the frozen vision transformer and concatenated with the encoded template-search tokens of the source domain images, allowing the model to rapidly find the optimal convergence checkpoints in various adverse weather conditions.

\subsection{Target-Aware Confidence Alignment}
\label{subsec:target-alignment}

Since the annotations are only available in the source domain, we train the tracker following a pseudo-label propagation strategy. Specifically, we send the synthesized template-search pairs into the teacher network to generate pseudo labels. These pseudo labels are fed back into the student network as supervision to update the weights of the tracking model. However, as the pseudo labels may be noisy, the incorrect pseudo labels will mislead the target state prediction. To address this problem, we propose a Target-Aware Confidence Alignment (TCA) module using optimal transport theory (OT) to enhance localization consistency in both domains by measuring the discrepancies in localization confidence at the candidate positions.

To be concrete, suppose the regressed response maps of student and teacher network are $\mathbf{r}^{\mathcal{S}} \in \mathbb{R}^{ N \times \left ( H' \times W' \right )  } $ and $\mathbf{r}^{\mathcal{T}}\in \mathbb{R}^{ N \times \left ( H' \times W' \right )  }$, where $N$ denotes the number of image samples in a mini-batch, $H'$, $W'$ represent the height and width of the response maps. We construct confidence distributions $\mathbf{d}^{\mathcal{S}} \in \mathbb{R}^N$ and $\mathbf{d}^{\mathcal{T}} \in \mathbb{R}^N$ for each sample in a mini-batch as:

\begin{equation}
    \mathbf{d}^{\mathcal{S}}= \exp(\mathbf{r}^{\mathcal{S}}_{i,\mathbf{p}_{i}}),\mathbf{d}^{\mathcal{T}}= \exp(\mathbf{r}^{\mathcal{T}}_{i,\mathbf{p}_{i}}),
\end{equation}
where for the $i$-th sample, $\mathbf{p}_{i} = \underset{j=1 \ldots H' \times W'}{\arg\max}
\mathbf{r}^{\mathcal{T}}_{i,j}$ denotes the spatial index of the response map with the
highest confidence score.

To construct the costmap $\mathbf{C}_{i,j}$ for the OT problem, we simultaneously consider the spatial and confidence discrepancies of each sample. Here we introduce two cost to measure the matching cost:

\vspace{-10pt}
\begin{equation}
    \mathbf{C}^{\textrm{Conf}}_{i,j}= \frac{\left \| \mathbf{r}^{\mathcal{S}}_{i,\mathbf{p}_{i}}-\mathbf{r}^{\mathcal{T}}_{j,\mathbf{p}_{j}} \right \|_{1} }{\max_{1\le m,n\le N }\left \| \mathbf{r}^{\mathcal{S}}_{m,\mathbf{p}_{m}}-\mathbf{r}^{\mathcal{T}}_{n,\mathbf{p}_{n}}  \right \|_{1}},
\end{equation}

\begin{equation}
    \mathbf{C}^{\textrm{Pos}}_{i,j}= \frac{\left \| \mathbf{p}^{\mathcal{S}}_{i}-\mathbf{p}^{\mathcal{T}}_{j} \right \|_{2} }{\max_{1\le m,n\le N }\left \| \mathbf{p}^{\mathcal{S}}_{m}-\mathbf{p}^{\mathcal{T}}_{n} \right \|_{2} },
\end{equation}

\begin{equation}
    \mathbf{C}_{i,j} = \mathbf{C}^{\textrm{Conf}}_{i,j} + \mathbf{C}^{\textrm{Pos}}_{i,j}
\end{equation}
where $\mathbf{C}^{\textrm{Conf}}$ and $\mathbf{C}^{\textrm{Pos}}$ represent the confidence and position cost between the distribution $\mathbf{d}^{\mathcal{S}}$ to $\mathbf{d}^{\mathcal{T}}$.

Based on what we discussed above, we design a \textbf{position-sensitive optimal transport (PSOT)} loss to measure the cost for moving the confidence distribution from $\mathbf{d}^{\mathcal{S}}$ to $\mathbf{d}^{\mathcal{T}}$, which can be  defined as the OT problem’s \textbf{dual} formulation:

\begin{equation}
    L_{p}= \left\langle \bm{\mu } ,\frac{\mathbf{d}^{\mathcal{T}}}{\left \| \mathbf{d}^{\mathcal{T}} \right \|_{1}}  \right \rangle +\left \langle \bm{\nu } ,\frac{\mathbf{d}^{\mathcal{S}}}{\left \| \mathbf{d}^{\mathcal{S}} \right \|_{1} }  \right \rangle.
\end{equation}
where  $\bm{\mu }$ and  $\bm{\nu }$ are the solutions of the OT problem. The details can be found in appendix B.





During training, we jointly adopt the target supervision loss and position-sensitive optimal transport loss as hybrid supervision loss to train the whole student-teacher model, which is given by:

\begin{equation}
\mathcal{L} = \mathcal{L}_t + \lambda \mathcal{L}_p,
\end{equation}
where $\lambda$ is the hyperparameter to balance the weights of the loss terms. We solve the OT problem by a fast Sinkhorn distances algorithm \cite{DBLP:conf/nips/Cuturi13}. Similar to \cite{ye2022joint}, the target supervision loss consists of the classification loss, localization $L_{1}$ loss and generalized GIoU loss as below:

\begin{equation}\label{stage1 loss}
L_{t} = \mathcal{L}_{cls} + \beta L_{1} + \gamma L_{GIoU}.
\end{equation}

By minimizing the target supervision loss and position-sensitive optimal transport loss, the feature representations and localization response can be effectively aligned to alleviate the domain shift.

\section{Experiments}

In this section, we conduct several experiments to evaluate the effectiveness of our proposed method. Our method is implemented based on python 3.10 and pytorch 2.1.1. Our tracker is trained with 4 NVIDIA RTX 3090 GPUs. All of the inference speed testing are conducted on a single NVIDIA RTX 3090 GPU.

\hspace*{-\dimexpr\tabcolsep} 
\begin{table*}[t]
\centering
\caption{Comparison with state-of-the-art visual trackers on synthetic datasets: GOT-10k-Foggy, DTB70-Foggy, GOT-10k-Dark, DTB70-Dark, GOT-10k-Rainy and DTB70-Rainy. The top two results are highlighted with \textcolor{red}{red} and \textcolor{blue}{blue} fonts, respectively. The double line above represents the \textbf{cross-domain}
trackers, while the line below represents the \textbf{generic} trackers.}
\label{tab:comparison}
\small
\setlength{\tabcolsep}{3pt} 
\vspace{-5pt}

\begin{tabular}{c|ccc|cc|ccc|cc|ccc|cc}
\toprule
\multirow{2.5}{*}{Tracker} & \multicolumn{3}{c|}{GOT-10k-Foggy} & \multicolumn{2}{c|}{DTB70-Foggy} & \multicolumn{3}{c|}{GOT-10k-Dark} & \multicolumn{2}{c|}{DTB70-Dark} & \multicolumn{3}{c|}{GOT-10k-Rainy} & \multicolumn{2}{c}{DTB70-Rainy} \\
\cmidrule(lr){2-16}
& AO & SR$_{0.50}$ & SR$_{0.75}$ & AUC & P & AO & SR$_{0.50}$ & SR$_{0.75}$ & AUC & P & AO & SR$_{0.50}$ & SR$_{0.75}$ & AUC & P \\
\midrule

\rowcolor{gray!20} \textbf{UMDATrack} & \textcolor{red}{66.6} & \textcolor{red}{75.8} & \textcolor{red}{62.2} & \textcolor{red}{66.21} & \textcolor{red}{86.05} & \textcolor{red}{65.4} & \textcolor{red}{75.3} & \textcolor{red}{57.3} & \textcolor{red}{66.07} & \textcolor{red}{85.72} & \textcolor{red}{68.5} & \textcolor{red}{78.4} & \textcolor{red}{63.2} & \textcolor{red}{66.75} & \textcolor{red}{87.60} \\

DCPT\cite{DBLP:conf/icra/ZhuTCH0QLL24} & 61.6 & 70.2
 & 56.9 & 58.31 & 75.33 & 62.4 & 70.5 & 54.2 & 61.87 & 80.11 & 62.3 & 70.1 & 59.8 & 61.68 & 82.56 \\

UDAT-CAR\cite{DBLP:conf/cvpr/Ye0ZPC22} & 51.5 & 60.3 & 45.2 & 50.21 & 69.41 & 56.8 & 64.2 & 49.1 & 57.20 & 75.80 & 59.5 & 65.2 & 55.3 & 56.42 & 75.36 \\

SAM-DA\cite{Yao2023SAMDA} & 50.2 & 60.5 & 48.3 & 51.33 & 69.89 & 55.4 & 63.1 & 48.3 & 57.15 & 75.12 & 60.2 & 66.1 & 57.6 & 57
63 & 76.12 \\

MLKD-Track\cite{DBLP:journals/corr/abs-2312-07884} & 52.3 & 62.3 & 49.1 & 52.46 & 70.32 & 53.8 & 61.6 & 46.9 & 55.21 & 73.68 & 57.3 & 64.8 & 57.1 & 56.89 & 74.12 \\

\midrule
\midrule

ARTrackV2\cite{Bai_2024_CVPR} & 64.8 & 73.0 & \textcolor{blue}{59.9} & \textcolor{blue}{62.25} & \textcolor{blue}{80.15} & \textcolor{blue}{63.1} & \textcolor{blue}{72.8} & 53.9 & 62.87 & 80.56 & \textcolor{blue}{66.2} & \textcolor{blue}{75.8} & 61.2 & 63.84 & 83.32 \\

EVPTrack\cite{shi2024evptrack} & 63.5 & 70.7 & 56.5 & 57.96 & 75.45 & 62.7 & 71.8 & 53.9 & \textcolor{blue}{63.01} & 81.12 & 65.5 & 75.2 & 60.5 & \textcolor{blue}{64.03} & \textcolor{blue}{84.11} \\

ODTrack\cite{zheng2024odtrack} & 65.1 & 74.5 & 56.0 & 61.12 & 79.32 & 62.5 & 71.5 & 53.1 & 62.21 & 80.23 & 64.8 & 74.5 & 59.5 & 63.95 & 83.56 \\

HipTrack\cite{cai2024hiptrack} & 63.3 & 72.0 & 59.6 & 60.52 & 78.22 & 62.9 & 72.4 & 53.8 & 62.48 & 80.57 & 65.6 & 75.4 & 60.2 & 63.57 & 83.36 \\

DropTrack\cite{dropmae2023} & 64.9 & 73.8 & 58.5 & 59.95 & 77.66 & 62.2 & 72.5 & \textcolor{blue}{54.3} & 61.98 & 80.21 & 65.3 & 75.3 & 60.4 & 62.87 & 83.13 \\

SeqTrack\cite{chen2023seqtrack} & \textcolor{blue}{65.2} & \textcolor{blue}{74.6} & 56.3 & 60.21 & 78.70 & 61.4 & 70.5 & 52.3 & 62.84 & \textcolor{blue}{81.57} & 65.1 & 75.0 & 60.3 & 63.75 & 83.28 \\

AQATrack\cite{xie2024autoregressive} & 64.9 & 72.8 & 59.7 & 57.28 & 75.61 & 61.7 & 70.6 & 52.5 & 61.17 & 79.87 & 63.4 & 72.3 & \textcolor{blue}{61.8} & 63.12 & 83.55 \\

ROMTrack\cite{Cai_2023_ICCV} & 63.6 & 70.9 & 56.7 & 59.05 & 76.59 & 60.8 & 71.1 & 51.7 & 60.80 & 77.95 & 62.7 & 73.4 & 60.1 & 63.21 & 83.25 \\

OSTrack\cite{ye2022joint} & 61.9 & 71.7 & 59.7 & 56.23 & 77.43 & 61.3 & 70.9 & 51.5 & 59.23 & 77.43 & 61.6 & 71.0 & 58.6 & 59.23 & 77.43 \\

AVTrack\cite{lilearning} & 56.9 & 63.5 & 49.5 & 52.35 & 68.09 & 55.3 & 62.3 & 46.2 & 56.66 & 72.21 & 57.5 & 63.4 & 48.1 & 60.21 & 79.53 \\

DiMP\cite{bhat2019learning} & 57.6 & 64.2 & 50.4 & 53.80 & 69.50 & 56.9 & 60.4 & 44.3 & 55.20 & 72.30 & 57.9 & 63.8 & 49.2 & 57.32 & 75.21 \\

SiamRPN++\cite{li2019siamrpn++} & 58.4 & 64.9 & 51.2 & 55.80 & 74.70 & 56.6 & 60.8 & 45.1 & 48.80 & 70.30 & 56.2 & 61.4 & 46.8 & 51.52 & 71.96 \\

SiamRPN\cite{li2018high} & 51.7 & 55.6 & 32.5 & 47.40 & 67.40 & 49.2 & 53.2 & 31.4 & 43.70 & 60.30 & 50.1 & 54.6 & 35.1 & 48.25 & 68.22 \\

\bottomrule
\end{tabular}
\vspace{-10pt}
\end{table*}

\vspace{-10pt}
\subsection{Implementation Details}

\noindent\textbf{Model settings.} We adopt vanilla ViT-Base \cite{DBLP:conf/iclr/DosovitskiyB0WZ21} model as the backbone of our tracker, similar to OSTrack\cite{ye2022joint}. The patch size is set to $16 \times 16$. We adopt a lightweight FCN consists of 4 stacked Conv-BN-ReLU layers as prediction head for both teacher and student branches. The sizes of the template and search region are resized to $128 \times 128$ and $256 \times 256$ respectively, corresponding to $2^2$ and $4^2$ times of the target box area.

\noindent\textbf{Training Details.} Our training process is divided into two stages: backbone training stage and domain customized training stage. We first synthesize the videos in adverse weather conditions only using GOT-10k dataset, the synthesized datasets includes GOT-10k-Dark, GOT-10k-Foggy and GOT-10k-Rainy. For backbone training, the DCA module is not introduced, we employ target supervision loss and position-sensitive optimal transport loss to perform domain adaptation between the teacher and student networks.
Four source domain datasets, including LaSOT \cite{DBLP:conf/cvpr/FanLYCDYBXLL19}, TrackingNet \cite{DBLP:conf/eccv/MullerBGAG18}, COCO \cite{coco}, and GOT-10k \cite{got10k}, as well as three synthetic datasets train the student model. The sampling ratio of the datasets is set to 1:1:1:1:4:4:4. The backbone training takes 250 epochs. The learning rate is $4\times 10^{-4}$ and decreased with weight decay $1 \times 10^{-4}$. The EMA hyperparameter $\alpha$ is set to 0.99. For domain customized training stage, we froze the backbone feature extractor and train the DCA module for an additional 50 epochs. Both two stages optimize the model with ADAMW. Note that our UMDATrack does not require repetitive backbone training stage, we only need to train the DCA module for each weather condition. Therefore, it only takes one and a half days to train UMDATrack in all weather conditions. This approach significantly improves training efficiency while maintaining superior model performance.


\noindent\textbf{Loss Function.} In our implementation, we utilize focal loss \cite{ross2017focal} for foreground-background classification and employ L1 loss and GIoU loss \cite{rezatofighi2019generalized,DBLP:journals/tip/YaoGYRC25}  for bounding box regression. Additionally, PSOT (Position-Sensitive Optimal Transport) loss is applied to align the distributions between the teacher and student networks. The weighting coefficients for the focal loss, L1 loss, GIoU loss, and PDOT loss are set to 1.0, 5.0, 2.0, and 10.0, respectively.

\noindent\textbf{Inference.} To accelerate the inference, the template feature is initialized using the first frame of each video sequence and stored for relation modeling between the template and search region in subsequent frames.
As demonstrated in Tab \ref{tab:comparison_fps}, we compared inference speed, MACs, and parameter counts with those of state-of-the-art trackers, showing that UMDATrack achieves the highest inference speed with relatively low computational costs and parameter counts.

\subsection{Comparisons with State-of-the-arts}
In this subsection, we comprehensively compare UMDATrack with SOTA trackers in both real-world and synthesized adverse weather conditions to demonstrate the effectiveness and high efficiency of our method. It's worth noting that our task is focused on cross-domain tracking, rather than being a generic one. However, we have observed significant performance improvement compared to the current state-of-the-art in generic trackers. 


Specifically, for nighttime conditions, we use the real-world NAT2021-test\cite{DBLP:conf/cvpr/Ye0ZPC22}, UAVDark70\cite{li2021adtrack}, and two synthesized datasets, i.e. GOT-10k-Dark, and DTB70-Dark. For foggy environment, we evaluate the tracking  performance using the GOT-10k-Foggy and DTB70-Foggy datasets. For rainy conditions, we use the GOT-10k-Rainy and DTB70-Rainy datasets. Finally, we use the real-world AVisT \cite{DBLP:conf/bmvc/NomanGK0DDC0GK22} dataset to evaluate the tracking performance under various adverse weather conditions in natural environment.

\noindent\textbf{Synthetic GOT-10k and DTB70\cite{drone-tracking}.}
As shown in Table \ref{tab:comparison}, UMDATrack performs exceptionally well across all three challenging conditions (foggy, dark, and rainy) on both the synthetic GOT-10k and DTB70 datasets.
Under dark conditions, UMDATrack achieved the highest AUC (66.07) and precision (85.72) on the DTB70-Dark dataset, outperforming the second-best resutls by a notable margin of 3.06\% in AUC and 4.15\% in precision. A similar trend is observed on the GOT-10k-Dark dataset, where UMDATrack leads both AUC and precision.
In foggy conditions, UMDATrack outperforms the second-best results obtained by other trackers by 3.96\% in AUC and 5.90\% in precision on the DTB70-Foggy dataset. In rainy conditions, UMDATrack also demonstrates superior performance to the advanced SOTA trackers. e.g. ARTrackV2 or ODTrack.

\begin{table}[]
\centering
\caption{Comparison with state-of-the-art visual trackers on real-world datasets: NAT2021, UAVDark70, and AVisT. The top two results are highlighted in \textcolor{red}{red} and \textcolor{blue}{blue}, respectively. The double line above represents the \textbf{cross-domain}
trackers, while the line below represents the \textbf{generic} trackers.}
\label{tab:comparison_nat}

\small
\setlength{\tabcolsep}{3.8pt}
\vspace{-5pt}

\begin{tabular}{c|cc|cc|cc}
\toprule
\multirow{2.5}{*}{Tracker} & \multicolumn{2}{c|}{NAT2021} & \multicolumn{2}{c|}{UAVDark70} & \multicolumn{2}{c}{AVisT}\\
\cmidrule(lr){2-7}
& AUC & P & AUC & P & AUC & P \\
\midrule
\rowcolor{gray!20} \textbf{UMDATrack} & \textcolor{red}{54.58} & \textcolor{red}{70.78} & \textcolor{red}{60.05} & \textcolor{red}{73.35} & \textcolor{red}{60.50} & \textcolor{red}{59.01} \\

DCPT\cite{DBLP:conf/icra/ZhuTCH0QLL24} & 52.55 & 69.01 & 56.86 & 70.16 & 55.66 & 52.41 \\

UDAT-CAR\cite{DBLP:conf/cvpr/Ye0ZPC22} & 48.75 & 65.96 & 51.25 & 70.22 & 38.91 & 33.65  \\

SAM-DA\cite{Yao2023SAMDA} & 47.31 & 65.50 & 49.52 & 65.59 & 37.36 & 34.29  \\

MLKD-Track\cite{DBLP:journals/corr/abs-2312-07884} & 44.31 & 60.21 & 47.27 & 61.54 & 33.62 & 30.26  \\

\midrule
\midrule

ARTrackV2\cite{Bai_2024_CVPR} & \textcolor{blue}{53.13} & \textcolor{blue}{69.72} & \textcolor{blue}{58.22} & \textcolor{blue}{71.95} & 58.52 & 57.65 \\

ODTrack\cite{zheng2024odtrack} & 53.11 & 69.68 & 58.07 & 71.11 & 58.63 & 57.36 \\

EVPTrack\cite{shi2024evptrack} & 53.08 & 69.51 & 57.47 & 71.10 & 57.31 & 55.55 \\

DropTrack\cite{dropmae2023} & 52.98 & 69.11 & 58.13 & 71.86 & \textcolor{blue}{59.56} & \textcolor{blue}{57.97} \\

ROMTrack\cite{Cai_2023_ICCV} & 51.57 & 68.75 & 53.77 & 69.80 & 56.12 & 55.09 \\

SeqTrack\cite{chen2023seqtrack} & 51.65 & 67.97 & 53.88 & 66.88 & 57.15 & 55.30 \\

AQATrack\cite{xie2024autoregressive} & 51.33 & 67.03 & 58.18 & 70.98 & 57.32 & 56.60 \\

SMAT\cite{gopal2024separable} & 45.96 & 59.87 & 45.19 & 56.71 & 50.35 & 49.58  \\

AVTrack\cite{lilearning} & 45.41 & 59.51 & 46.91 & 59.49 & 49.21 & 48.50 \\

\bottomrule
\end{tabular}
\vspace{-10pt}
\end{table}

\begin{table}[]
\centering
\caption{Comparison of inference speed, FLOPs, and model parameters across different trackers.}
\label{tab:comparison_fps}
\small
\vspace{-5pt}

\begin{tabular}{c|c|c|c} 
\toprule
Tracker & Speed (FPS) & MACs (G) & Params (M) \\ 
\midrule

\rowcolor{gray!20} \textbf{UMDATrack} & \textbf{138} & \textbf{18} & \textbf{65} \\

ARTrackV2\cite{Bai_2024_CVPR} & 95  & 45 & 126  \\

EVPTrack\cite{shi2024evptrack} & 71 & 22 & 74 \\

AQATrack-\cite{xie2024autoregressive} & 68 & 26 & 72 \\

DropTrack\cite{dropmae2023} & 52 & 48 & 92 \\

SeqTrack\cite{chen2023seqtrack} & 40 & 66 & 89 \\

\bottomrule
\end{tabular}
\vspace{-10pt}
\end{table}

\noindent\textbf{Results on Real-World datasets}
To further verify the effectiveness of the proposed UMDATrack, we conduct experiments on the real-world datasets with adverse weather conditions for comparison. As shown in Table \ref{tab:comparison_nat}, on the large-scale night dataset NAT2021, UMDATrack achieved the best AUC (54.58) and precision (70.78). Specifically, in terms of AUC, we outperformed the second tracker ARTrackV2 (53.13) by 1.45 points. This partially proves that our proposed framework helps the model learn effectively from synthetic extreme domain datasets. For the challenging UAV tracking dataset UAVDark70, UMDATrack outperforms all other trackers on the UAVDark70 real-world dataset, achieving an AUC score 1.83 points higher and a precision 1.4 points greater than the second-best tracker. Note that most of the reported trackers in the table can not directly deployed run for UAV system. However, UMDATrack obtains the best performance with real-time speed, shown great potential in real-world UAV tracking. Furthermore, we also test UMDATrack on AVisT dataset, which is specifically collected for tracking in diverse scenarios
with adverse visibility. The various weather conditions such as rain, snow, fog and camouflage are included in this dataset, UMDATrack also obtains the leading performance in both precision and AUC metrics.

\noindent\textbf{Inference Speed.} Since UMDATrack does not require to introduce heavy blocks for target appearance model, the computational cost of UMDATrack is limited. As demonstrated in Table \ref{tab:comparison_fps}, we compared inference speed, MACs, and parameter counts with those of state-of-the-art trackers, showing that UMDATrack achieves the highest inference speed with relatively low computational costs and parameter counts.

\subsection{Ablation Studies and Visualization}

\begin{table}[]
\centering
\caption{Ablation study on the individual impact of each module (CSG, DCA, and TCA) in our model. The presence or absence of each module is marked with a check or dash, respectively. Results are reported in terms of AUC and Precision for each configuration, evaluated on the NAT2021 dataset.}
\label{tab:effectiveness_metrics}

\small

\begin{tabular}{@{}ccc|cc@{}} 
\toprule
\multicolumn{3}{c|}{\textbf{Modules}} & \multicolumn{2}{c}{\textbf{Indicators}} \\  
\midrule
CSG & DCA & TCA & AUC (\%) & Precision (\%) \\
\midrule
- & - & - & 49.11 & 63.52 \\
$\checkmark$ & - & - & 50.90 & 65.38 \\
- & $\checkmark$ & - & 50.56 & 65.50 \\

$\checkmark$ & - & $\checkmark$ & 52.27 & 67.10 \\
$\checkmark$ & $\checkmark$ & - & 52.24  & 67.49 \\
$\checkmark$ & $\checkmark$ & $\checkmark$ & \textbf{54.58} & \textbf{70.78}  \\
\bottomrule
\end{tabular}
\vspace{-15pt}
\end{table}

\noindent\textbf{Study on the components of UMDATrack.} We conducted ablation experiments on the proposed three modules to verify their effectiveness. As shown in Table \ref{tab:effectiveness_metrics}, the baseline approach doesn't introduce any modules, thus it is only trained only on the four source domain datasets. When the CTG module is introduced, the model achieves the AUC of 50.90\% and Precision of 65.38\%. Adding the TCA module improves these results, bringing the AUC to 52.27\% and precision to 67.10\%. Further including the DCA module increases performance to the AUC of 54.58\% and Precision of 70.78\%. These results demonstrate that each module provides a significant performance gain, with the full model configuration yielding the highest scores in both metrics on the NAT2021 dataset.

\begin{table}[H]
\centering
\caption{Effect of different EMA (Exponential Moving Average) update frequencies on model performance.}
\label{tab:ema_frequency}
\small
\vspace{-5pt}

\begin{tabular}{@{}c|cc@{}}
\toprule
EMA Frequency  & AUC (\%)  & Precision (\%) \\
\midrule
Each epoch      & \textbf{54.48} & \textbf{70.78} \\
Every 3 epochs  & 53.65 & 68.99 \\
Every 5 epochs  & 52.90 & 68.22 \\
Each batch      & 53.89 & 69.57 \\
\bottomrule
\end{tabular}
\vspace{-25pt}
\end{table}

\begin{table}[H]
\centering
\caption{Different dataset proportions used for training, with LaSOT, GOT-10k, TrackingNet, COCO, and Synthetic datasets in the specified ratios.}
\label{tab:dataset_proportion}
\small
\vspace{-5pt}

\begin{tabular}{@{}c|cc@{}}
\toprule
Dataset Proportion  & AUC (\%) & Precision (\%) \\
\midrule
1 : 1 : 1 : 1 : 1 : 1 : 1   & 53.13 & 68.72 \\
1 : 1 : 1 : 1 : 2 : 2 : 2   & 53.68 & 69.01 \\
1 : 1 : 1 : 1 : 4 : 4 : 4   & \textbf{54.58} & \textbf{70.78} \\
1 : 1 : 1 : 1 : 6 : 6 : 6   & 54.26 & 70.44 \\

\bottomrule
\end{tabular}
\vspace{-10pt}
\end{table}

\noindent\textbf{Study on the training hyper-parameter of UMDATrack.}
We conducted two ablation studies on the update frequency of EMA and the proportion of the training dataset.
As shown in Table \ref{tab:ema_frequency}, we experimented with performing EMA after
each epoch, every three epochs, every five epochs, and after completing each batch to transfer student network's weight to the teacher network. The results indicate that performing EMA after each epoch yields the best results.
For the dataset proportion settings, we conducted four groups of experiments as shown in the Table \ref{tab:dataset_proportion}, and the results indicate that group 3 achieve the best performance. Therefore, we set the training dataset proportion to 1:1:1:1:4:4:4.

\begin{figure}[ht]
\includegraphics[width=3.3in]{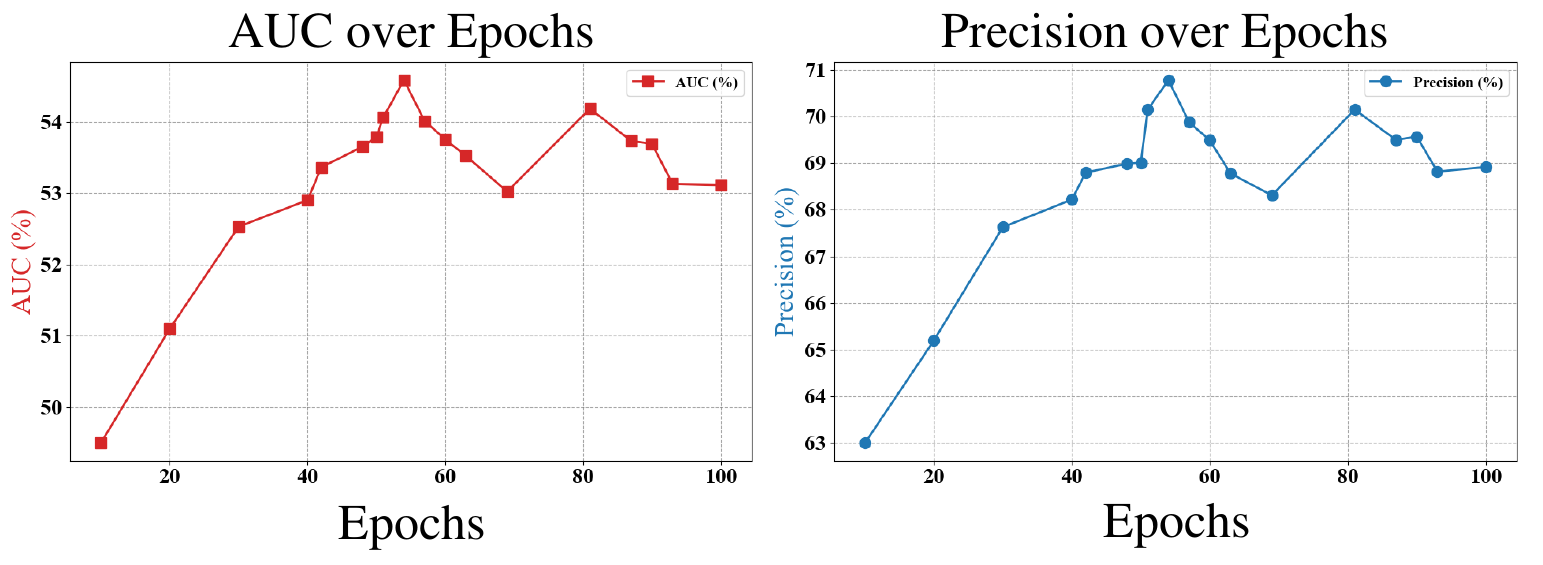}

\caption{The convergence speed of DCA. Please zoom in for details.}
\label{fig:adapter_convergence}
\vspace{-10pt}
\end{figure}

\begin{figure}[htbp]
  \centering
  \includegraphics[width=0.5\textwidth]{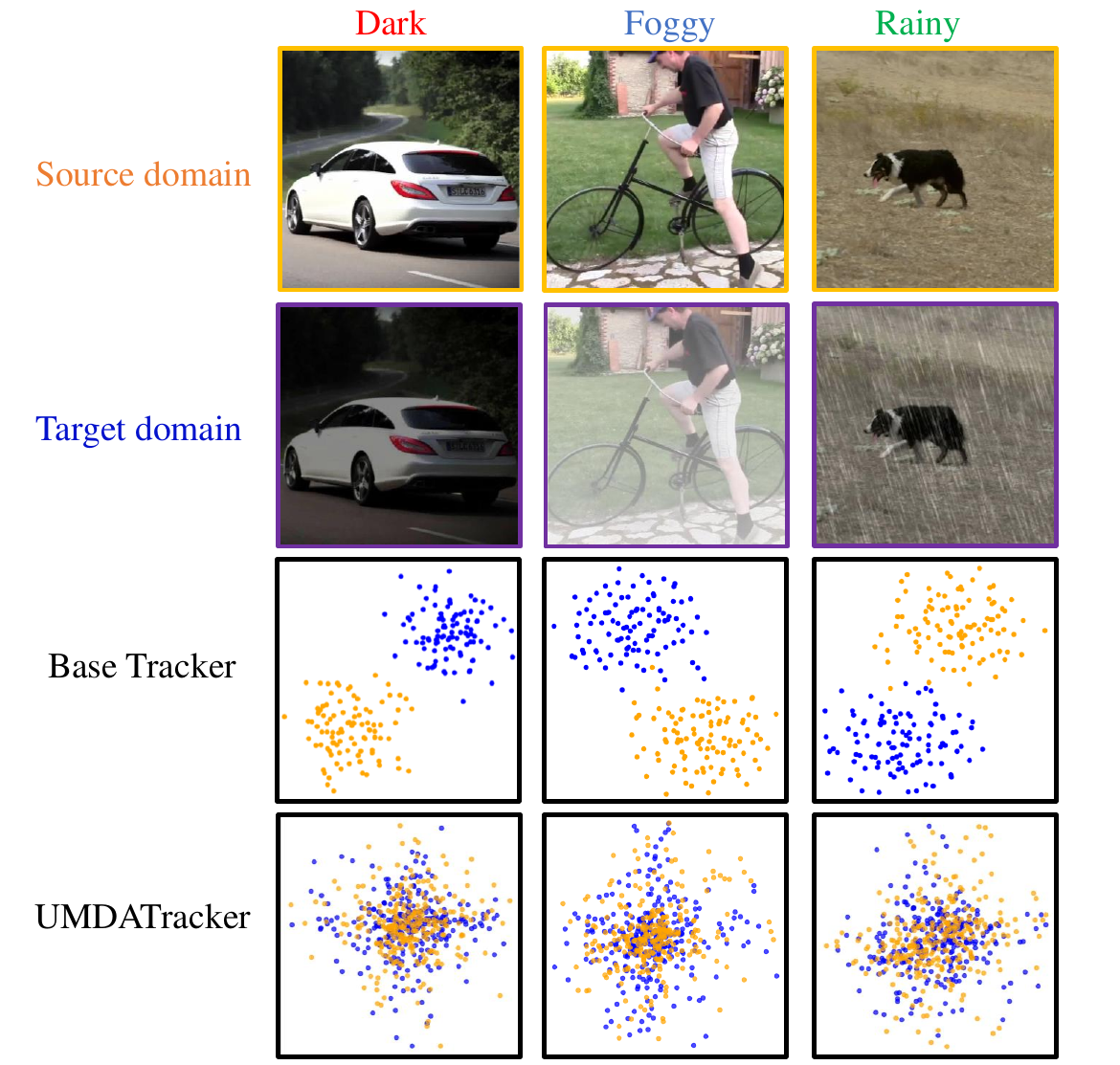}

   \caption{Feature visualization by t-SNE of dark, foggy, and rainy scenes compared to normal (daytime) scenes. \textcolor[HTML]{f9a718}{Orange} and 
   \textcolor{blue}{blue} 
    indicate source domain and target domains, respectively. The scattergrams depict the feature distributions of the base tracker and UMDATracker across different weather conditions. The results show that UMDATracker effectively narrows the domain discrepancy in various challenging weather conditions.}

  \label{fig:tsne}
\end{figure}



\vspace{-20pt}

\begin{table}[H]
\centering
\caption{Quality comparison of synthetic datasets generated by different generators. AUC is evaluated on NAT2021 dataset.}
\label{tab:different_generator}
\small

\setlength{\tabcolsep}{1pt}
\vspace{-5pt}

\begin{tabular}{c|cc|c|c}
\toprule
Method & SSIM$\uparrow$ & LPIPS$\downarrow$ & Time (h) & AUC (\%) \\
\midrule
\rowcolor{gray!20} \textbf{CSG with Text} & \textbf{0.920} & \textbf{0.086} & \textbf{24} & \textbf{54.58} \\
CSG without Text  & 0.902 & 0.104 & 20 & 52.53 \\
CycleGAN \cite{CycleGAN2017} & 0.895 & 0.119 & 30 & 51.10 \\
UNIT\cite{liu2017unsupervised} & 0.875 & 0.136 & 14 & 50.23 \\
Gamma(only for dark) & 0.787 & 0.216 & 5 & -  \\

\bottomrule
\end{tabular}
\vspace{-10pt}
\end{table}

\noindent\textbf{Study on the speed of DCA convergence.} We analyze the convergence speed in which the DCA achieves its optimal performance during training. As shown in Fig. \ref{fig:adapter_convergence}, around 50 epochs, the DCA has already obtained encouraging performance. Beyond this point, performance increases only slightly, and may even decline with additional epochs. Therefore, we suggest a trade-off between performance and training time to achieve efficiency.

\noindent\textbf{Study on the impact of the synthetic datasets.}
We use SSIM \cite{DBLP:journals/tip/WangBSS04} and LPIPS\cite{zhang2018perceptual} to evaluate image quality in the second and third columns of Table \ref{tab:different_generator}, Compared to other methods like CycleGAN, UNIT, or simply using Gamma,
CSG especially with text prompt achieves the best generation quality.
Although our generator requires slightly more time to synthesize datasets, this is a trade-off between data generation quality and computational time. The use of text prompts improves the quality and relevance of the generated datasets, leading to better downstream performance. As a result, the tracker achieves the best AUC performance.



\begin{figure}[]
\hspace*{-1em} 
\includegraphics[width=0.5\textwidth]{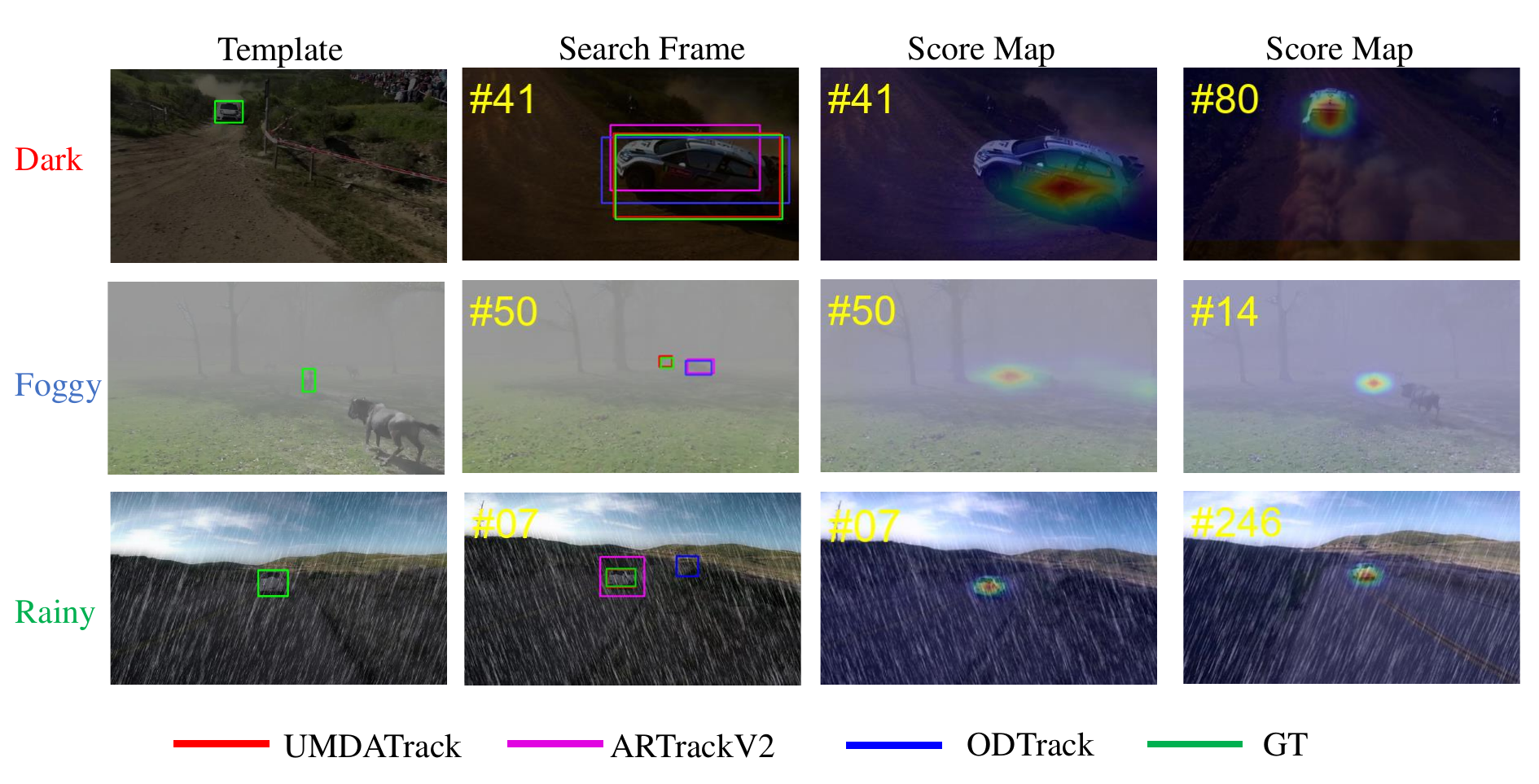}
\caption{Visualization comparison of  our approach and other excellent trackers and results of the scoremaps.}

\label{fig:bbox}
\vspace{-10pt}

\end{figure}

\noindent\textbf{Visualizing Robustness in Adverse Conditions.}
Fig. \ref{fig:tsne} shows feature distributions using t-SNE\cite{hinton2008visualizing}, where UMDATrack better aligns source domain and target domain across dark, foggy, and rainy conditions, reducing domain discrepancy.
Fig. \ref{fig:bbox} presents tracking results, with UMDATrack achieving higher accuracy and significantly stronger resistance compared to other trackers  in extreme scenarios.

\section{Conclusion}
\label{sec:conclusion}

In this paper, we propose a unified multi-domain adaptive tracker termed UMDATrack to predict target state under various adverse weather conditions. We first use a controllable scenario generator to synthesize unlabeled videos in multiple weather conditions under the guidance of different text prompts. Afterwards, we propose a simple yet effective domain-customized adapter to remedy the tracking model, allowing it to rapidly adapt to various weather conditions without redundant model updating. Furthermore, we propose a target-aware confidence alignment module (TCA) with optimal transport theorem, which enhances the localization consistency between source and target domains by measuring the discrepancies of the localization confidence at the candidate positions. Experiments show that UMDATrack leads new state-of-the-art performance on either real-world or synthesized datasets by a significant margin.

{\small
\bibliographystyle{ieee_fullname}
\bibliography{main}
}


\end{document}